\documentclass{article}

\usepackage{arxiv}

\usepackage[utf8]{inputenc} % allow utf-8 input
\usepackage[T1]{fontenc}    % use 8-bit T1 fonts
\usepackage{hyperref}       % hyperlinks
\usepackage{url}            % simple URL typesetting
\usepackage{booktabs}       % professional-quality tables
\usepackage{amsfonts}       % blackboard math symbols
\usepackage{nicefrac}       % compact symbols for 1/2, etc.
\usepackage{microtype}      % microtypography
\usepackage{lipsum}
\usepackage{graphicx}
\graphicspath{ {./images/} }
\usepackage{todonotes}
\usepackage[acronym]{glossaries}
\usepackage{caption}
\usepackage{subcaption}
\usepackage{amsmath}
\usepackage{amsfonts}
\usepackage{amssymb}
\usepackage{float}
\usepackage{alphabeta}
\usepackage{CJKutf8}
\usepackage{graphicx}
\usepackage{enumerate}
\usepackage{xcolor}
\usepackage{natbib}

\usepackage{prodint}
\usepackage[printonlyused,withpage]{acronym}
\usepackage[titletoc]{appendix}
\usepackage{hyperref}
\usepackage{chemfig}
\usepackage{longtable}
\usepackage{emptypage}
\usepackage{lscape}
\usepackage{longtable}
\usepackage{color, colortbl}
\makeglossaries
\newacronym{atm}{ATM}{Air Traffic Management}
\newacronym{pmm}{PMM}{Point-Mass Models}
\newacronym{bnn}{BNN}{Bayesian Neural Network}
\newacronym{dl}{DL}{Deep Learning}
\newacronym{ml}{ML}{Machine Learning}
\newacronym{lstm}{LSTM}{Long-Short Term Memory}
\newacronym{gan}{GAN}{Generative Adversarial Networks}
\newacronym{mdn}{MDN}{Mixture Density Networks}
\newacronym{gmm}{GMM}{Gaussian Mixture Models}
\newacronym{3dgmm}{3-D GMM}{3-D Gaussian Mixture Models}
\newacronym{mape}{MAPE}{Mean Absolute Percentage Error}
\newacronym{sem}{SEM}{Standard Error Mean}
\newacronym{ig}{IG}{Integrated Gradient}
\newacronym{vg}{VG}{Vanilla Gradient}
\newacronym{xai}{XAI}{EXplainable Artificial Intelligence}

\title{Handling Weather Uncertainty in Air Traffic Prediction through an Inverse Approach}

\author{
 G. Lancia \\
  Department of Basic and Applied Sciences for Engineering (SBAI)\\
  University of Rome "La Sapienza"\\
  \texttt{giacomo.lancia@uniroma1.it} \\
  \AND
  D. Falanga \\
  Department of Computer Science, Systems, and Communication\\
  University of Milano-Bicocca\\
  \texttt{davide.falanga@unimib.it} 
 \AND
  S. Alam \\
School of Mechanical \& Aerospace Engineering \\  
Nanyang Technological University, Singapore \\  
\texttt{sameer.alam@ntu.edu.sg}
\AND
  G. Lulli \\
  Department of Computer Science, Systems, and Communication\\
  University of Milano-Bicocca\\
    \texttt{guglielmo.lulli@unimib.it} 
}

\begin{document}

\maketitle
\begin{abstract}

Adverse weather conditions, particularly convective phenomena, pose significant challenges to Air Traffic Management, often requiring real-time rerouting decisions that impact efficiency and safety. This study introduces a 3-D Gaussian Mixture Model to predict long lead-time flight trajectory changes, incorporating comprehensive weather and traffic data. 
Utilizing high-resolution meteorological datasets, including convective weather maps and wind data, alongside traffic records, the model demonstrates robust performance in forecasting reroutes up to 60 minutes. 
The novel 3-D Gaussian Mixture Model framework employs a probabilistic approach to capture uncertainty while providing accurate forecasts of altitude, latitude, and longitude. 
Extensive evaluation revealed a Mean Absolute Percentage Error below 0.02 across varying lead times, highlighting the model's accuracy and scalability. 
By integrating explainability techniques such as the Vanilla Gradient algorithm, the study provides insights into feature contributions, showing that they contribute to improving Air Traffic Management strategies to mitigate weather-induced disruptions.

\end{abstract}
\newpage
\section{Introduction}

The exponential growth in global air traffic, combined with increasingly unpredictable weather patterns due to climate change, has underscored the critical need for robust, real-time flight trajectory prediction systems \cite[]{reitmann2019advanced, kontogiannis2017cognitive}. 
\gls{atm} aims to ensure the safety and efficiency of air operations by reducing delays, mitigating potential conflicts, and managing environmental impacts \cite[]{lulli2007european}. 
One of the most pressing challenges in this domain is predicting flight route deviations in response to adverse weather conditions, particularly convective weather, which is responsible for significant disruptions in air traffic operations. 
Over the years, adverse weather has been one of the most disruptive factors for \gls{atm} in Europe, emerging as a major source of logistical challenges and increased costs \cite[]{cook2011european, eurocontrol2023}.
Among various weather-related causes, convective weather, characterized by cumulonimbus clouds, thunderstorms, and severe turbulence, presents substantial hazards to aviation \cite[]{wmo_aviation2023}, accounting for 30 of en-route delays in Europe alone \cite{eurocontrol2023}.

This paper primarily focuses on predicting long lead-time trajectory changes, with significant emphasis on modeling these changes by incorporating weather-based information.
In recent years, physics-based models, such as \gls{pmm}, have become a common choice to predict aircraft trajectories by leveraging aerodynamic and performance parameters \cite[]{zhang2018online}.
However, this approach limits the integration of dynamic weather conditions, making it difficult to adapt them for comprehensive and real-time forecasting.
Among data-driven approaches, \gls{bnn} and recurrent architectures have significantly improved trajectory prediction accuracy \cite[]{zhang2020bayesian}.
The existing models, however, may be lacking in handling real-time weather-induced deviations. 

The importance of addressing these challenges is further accentuated by the growing airspace demand and the increasing intensity of weather disturbances globally. 
As suggested by \cite{radler2019frequency}, the frequency of severe thunderstorms is expected to rise in the coming decades, putting further strain on the \gls{atm} system. 
Developing models capable of anticipating reroutes in such dynamic environments is crucial for reducing delays, improving fuel efficiency, and ensuring passenger safety.

The recent advancements of  \gls{ml} and \gls{dl} across various fields have also revolutionized trajectory prediction within the context of \gls{atm}.
Methods such as \gls{lstm} networks and \gls{gan} have shown promise in modelling sequential data and simulating possible future flight trajectories \cite[]{shi20204, zhu2024predicting, pang2020conditional}. 
In addition, recent research has highlighted the potential of \gls{mdn}, particularly \gls{gmm}, to represent a valuable approach to model uncertainty in flight trajectory predictions by capturing the probability distributions of future positions \cite[]{chen2020predicting}. 
Unlike the aforementioned approaches, such as \gls{lstm} and \gls{gan}, the \gls{mdn} models are based on an inverse problem strategy, allowing for a major control on the uncertainty of predictions \cite[]{herzallah2004mixture, herzallah2003multi} along with a flexible handling of the input data.
In an inverse problem approach, the goal is to estimate the underlying causes, such as the trajectory adjustments, given observed effects, like adverse weather conditions, thereby allowing the model to explicitly learn the mapping from consequences back to potential influencing factors.
As a result, \gls{mdn} represents a promising forecasting tool for \gls{atm} when integrating meteorological data.

Therefore, this paper seeks to advance current trajectory prediction models by introducing a novel \gls{gmm} designed to predict flight reroute decisions well in advance, accounting for various weather conditions. 
Unlike previous studies cited, which might be lacking in considering weather data or can often focus on a single aspect of weather-induced disruptions, our approach leverages a broader collection of meteorological datasets (such as wind, temperature, and convective weather data) along with historical flight trajectories. 
By modelling route changes as probabilistic events, we seek to embed weather-induced disruptions within a robust probabilistic framework. 
This integration is designed to enhance both the accuracy and reliability of long-range trajectory predictions, allowing for better-informed decision-making in \gls{atm}. 
Utilizing this probabilistic approach not only addresses the inherent uncertainties of weather impacts on flight paths but also facilitates adaptive strategies that can improve operational efficiency and safety in aviation

The primary objective of this research is to develop a model that can predict reroutes up to 60 minutes, ideally giving air traffic controllers ample time to make strategic decisions as soon as risky scenarios are highly likely to occur.
Using a combination of real-time flight data and weather forecasts, we aim to inspect the effects of various weather scenarios on flight paths, reducing the cognitive load on traffic managers and improving the overall resilience of the \gls{atm} system.

In summary, our work has the scope of contributing to the field of \gls{atm} by integrating \gls{gmm} by improving the accuracy of trajectory predictions under adverse weather conditions. 
By doing so, we hope to enhance air traffic management systems’ ability to mitigate weather-related disruptions and contribute to safer and more efficient air operations globally.

\section{Data}

%%%Data acquisition
\subsection{Data acquisition}
%1. Introduce the data. OpenSky
To conduct this study, we collected a vast air traffic dataset from \cite{OpenskyNetwork}, which provides high-resolution ADS-B Out data in Mode-S format. 
In specific, we employed 10-second resolution air traffic data.
The dataset was selected with the goal of acquiring a comprehensive coverage of relevant factors, such as the \emph{position} of each aircraft, their \emph{ground speed}, and other features like the \emph{heading} and the \emph{vertical rate} 
%2. ECW
Alongside this, we also gathered historical weather data from \cite{EumetsatConvectiveWeather}, with a special focus on \emph{convective weather}.
Among all available weather data, this dataset offers an interesting insight into the precipitation maps generated by combining infrared (IR) imagery from geostationary (GEO) satellites with calibrated precipitation measurements from microwave (MW) sensors on Low Earth Orbit (LEO) satellites. 
In addition, the \emph{"Rapid Update"} algorithm used to generate the precipitation maps blends $10.8 \mu m$ equivalent blackbody temperatures (TBB) from GEO IR images with rain rates derived from MW measurements. Convective precipitation detection is further refined using NEFODINA software, which employs morphological analysis to enhance precipitation estimates, particularly for intense, localized rainfall. 
This approach allows for high-frequency, near-real-time precipitation mapping, improving both accuracy and temporal resolution for effective weather monitoring and early-warning applications. 
The data are provided with a 15-minute temporal resolution.
To strengthen our analysis, we accounted for another influencing air traffic flow such as atmospheric airflow dynamics.
To do so, we acquired data from \cite{ERA5dataset}, available via the Copernicus Climate Data Store.
%3. ERA5
Note that \cite{ERA5dataset} is a fifth-generation ECMWF reanalysis, providing high-resolution weather and climate data from 1940 to the present.
Specifically, we focused on the \emph{u-} (east-west direction) and \emph{v-components} (north-south direction) of wind direction within a specific geographic area of interest.
With hourly data and pressure-level attributes, ERA5 allows us to analyze wind conditions at various altitudes, which are essential for assessing atmospheric impacts on air traffic. 
Specifically, we narrowed our focus on wind data specific for a common choice of altitude such as 38,000 feet.
%4. Recap the complete dataset
The complete dataset, therefore, consists of data acquired from three different sources: air traffic data, convective weather data, and wind data; see Table \ref{tab: complete_dataset} 
We restricted our attention to the Maastricht Upper Area Control Centre (MUAC) and acquired data from this highly active and complex airspace, one of the busiest European traffic regions.
To have an adequate and assorted collection of weather scenarios, we opted to acquire all available data over all days of January and May 2024.
These two months were selected based on an analysis of weather images, specifically examining pixel intensity and colour to determine weather severity. 
%Specifically, we chose January as a month with low levels of adverse weather and May as a month with frequent adverse weather conditions.
Specifically, this choice was based on a progressive pixel-wise analysis, ranging from light to dark colours. 
Under this assumption that in the absence of adverse weather actual flights closely follow flight plans with minimal deviations, we gathered the corresponding air traffic and wind data.
% Data Information. Table
\begin{table}[]
    \centering
    \begin{tabular}{|c|c|c|c|}
    \hline
         \textbf{Feature} &  \textbf{Data Source} & \textbf{Time Resolution} & \textbf{Data Type}\\
         \hline
         Convective Weather & \cite{EumetsatConvectiveWeather} & 15 minutes & Images\\
         u-component & \cite{ERA5dataset} & 60 minutes & Images\\
         v-component & \cite{ERA5dataset} & 60 minutes & Images\\
         Latitude & \cite{OpenskyNetwork} & 10 minutes & Array\\
         Longitude & \cite{OpenskyNetwork} & 10 minutes & Array\\
         Altitude & \cite{OpenskyNetwork} & 10 minutes & Array\\
         Ground Speed & \cite{OpenskyNetwork} & 10 minutes & Array\\
         Heading & \cite{OpenskyNetwork} & 10 minutes & Array\\
         Vertical Rate & \cite{OpenskyNetwork} & 10 minutes & Array\\  
         \hline
    \end{tabular}
    \caption{Summary description of the complete traffic dataset}
    \label{tab: complete_dataset}
\end{table}
%%%Dataset Creation. It needs a separated subsection
\subsection{Dataset Creation}
The instances in our dataset were constructed through a multi-step process to ensure relevance and accuracy for the study objectives. 
The selection of instances depends on the chosen lead time, as this dictates the prediction horizon and uniquely determines the instance creation process.
To begin, we established a set of lead times for analysis, selecting intervals of 1, 2, 5, 10, 30, 45, and 60 minutes for convenience.
Next, the traffic data are taken into account, setting a 1-minute sampling interval instead of the original 10-second interval.
Each instance represents the collection of traffic variables for a specific flight, where data is available at a given time $t$ and at $t + \tau$ ($\tau$ denotes the lead time).
Alongside the traffic data, each instance is equipped with both the convective weather and wind data from the first available time prior to $t$.
Recall that both the ERA5 and EUMETSAT data show a sampling frequency higher than the one of the OpenSky data.
Thus, we supplied each instance with the most recent available information about the convective weather and the wind flows.
As a response (target) variable, we opted for the position of an aircraft after the lead time $\tau$; by position we mean a 3-D array including coordinates like altitude, latitude, and longitude.
This process resulted in constructing the final dataset of about five millions instances. 
Further pre-processing operations will be discussed in detail in section \ref{sec: pre-processing}.

\section{Methodology}
To investigate long lead-time flight position forecasting with the integration of weather-based data, we implemented a \gls{mdn}-based approach.
This \gls{dl}-based technique was initially proposed in \cite{bishop1994mixture}; see also \cite{bishop2006pattern}.
When analyzing one-dimensional historical data, such a modelling technique has demonstrated significant success in effectively capturing complex non-linear dynamics grounded in physical principles. 
Recent studies, such as those by \cite{petersik2020probabilistic, lancia2022physics} have highlighted the ability of \gls{mdn} to capture intricate dynamics in non-linear climate physics-based models.

Within the context of flight trajectory prediction, we have readapted a \gls{gmm}-based prediction model for developing dynamic 3-D forecasts of altitude, longitude, and latitude at specified future times based on historical information.
From a modelling point of view, such a 3-D modelling forecast represents a novelty, since a vast majority of applications of \gls{gmm} are based on predicting one-dimensional variables.
In the following, we shall introduce the prediction model we developed in more detail, i.e., the \gls{3dgmm}.

\subsection{Model's Mathematical Foundations}
Similarly to the 1-D \gls{gmm}, the \gls{3dgmm} is focused on solving a probabilistic regression.
In our case, we attempted to solve a 3-D probabilistic regression. 
Therefore, the \gls{3dgmm} consists of learning the parameters of a 3-D mixture of normal distributions modelling the probability that a flight will be observed at a specific point (in terms of altitude, longitude, and latitude) given the current flight information.

Let us suppose $X_{i}(\tau)$ to be a 3-dimensional array containing the longitude, the latitude, and the altitude of the generic $i$-th flight at time $\tau$.
For convenience, we consider a generic neighbourhood of $X_{i}(\tau)$ and denote with $W_{i}(\tau)$ the 2-d grid-structured features such as the \emph{convective weather} and both the \emph{u-} and \emph{v-components}; with $\xi_{i}(\tau)$ we denote to the \emph{traffic features}. 
The weather features therefore refer to the weather scenario at time $\tau$, while the traffic features encompass both the baseline and other time-dependent flight variables, providing a complete description of the flight under consideration. 
Thus, the \gls{3dgmm} aims to learn the parameters ruling the desired mixture of $N$ generic 3-D normal distributions expressing the position of a flight at later times, namely
\begin{equation}
\begin{split}
    \mathbf{P}(X_{i}(\tau+\Delta\tau) |& W_i(\tau), \xi_{i}(\tau)) = \\
  =  \sum_{k = 1}^{N} & \frac{\alpha_{k}(W_i, \xi_i)}{\sqrt{(2\pi)^{3}|\Sigma_{k}|}} \exp\left(-\frac{1}{2} [X_i(\tau+\Delta\tau)-\mu_{k}(W_i, \xi_i)]^{T}\Sigma_{k}^{-1}(W_i, \xi_i) [X_i(\tau+\Delta\tau)-\mu_{k}(W_i, \xi_i)]\right)
\end{split}
\end{equation}
for any generic interval of time $\Delta \tau.$
Note that $\Delta \tau$ is the lead time of predictions. Given the information at time $\tau$, we will estimate how likely a flight will be observed on a specific area of the flying space with a lead time of $\Delta \tau.$
We recall that $\mu_{k}$ is the mean vector of positions for the $k$-th multivariate normal distribution and
$\Sigma_k$ is the symmetric positive-valued covariance matrix of the $k$-th multivariate normal distribution.
The coefficient $\alpha_{k}$ refers to a real positive-valued quantity.
To ensure all events sum up to unity, it is necessary to meet the constraint
\begin{equation}\label{eq: condition_prob_sum_1}
    \sum_{k = 1}^N \alpha_{k}(W_i, \xi_i) = 1.
\end{equation}
Note that this condition must be satisfied for any flight $i.$

As mentioned, the \gls{3dgmm} aims at modelling how likely the flight $i$-th will appear in the position $X_{i}(\tau + \Delta\tau)$ given the flight information at time $\tau.$  
Therefore, it is natural to consider as \emph{loss function} the log-likelihood function of all occurrences (say, $M$), namely
\begin{equation}\label{eq: Loss_function}
\begin{split}
  &  \Lambda(\alpha_1, \mu_1, \Sigma_1, \dots, \alpha_N, \mu_N, \Sigma_N) = \\ & = \frac{1}{M} \sum_{i = 1}^{M}  \log{\left[ \sum_{k = 1}^{N} \frac{\alpha_{k}(W_i, \xi_i)}{\sqrt{(2\pi)^3|\Sigma_k|}} \exp\left(-\frac{1}{2} [A_i-\mu_{k}(W_i, \xi_i)]^{T}\Sigma_{k}^{-1}(W_i, \xi_i) [A_i-\mu_{k}(W_i, \xi_i)]\right)\right]}. 
\end{split}
\end{equation}
By minimizing \eqref{eq: Loss_function}, we get the set of optimal parameters $\{\hat{\alpha}_1, \hat{\mu}_1, \hat{\Sigma}_1, \dots, \hat{\alpha}_N, \hat{\mu}_N, \hat{\Sigma}_N\}.$

%How to make predictions
As a method to approach inverse problems, the \gls{3dgmm} does not directly provide predictions.
However, the knowledge of the distribution parameters allows us to formulate predictions.
In particular, we focused on the scenario with the maximal occurrence.
Thus, the predicted position $\hat{X}_{i}(\tau + \Delta\tau)$ is evaluated through the formula
\begin{align}\notag
    \hat{X}_{i}(\tau + \Delta\tau) = \operatorname{argmax}_{X_{i}(\tau+\Delta\tau) \in \mathcal{F}} \Bigg\{ 
    \sum_{k = 1}^{N} 
    \frac{\hat{\alpha}_{k}(W_i, \xi_i)}{\sqrt{(2\pi)^{3}|\hat{\Sigma}_{k}|}} 
    \exp\Bigg(-\frac{1}{2} \big[X_i(\tau+\Delta\tau)-\mu_{k}(W_i, \xi_i)\big]^{T} \\
    \hat{\Sigma}_{k}^{-1}(W_i, \xi_i) 
    \big[X_i(\tau+\Delta\tau)-\mu_{k}(W_i, \xi_i)\big] 
    \Bigg) 
    \Bigg\}.
\end{align}
with $\mathcal{F}$ the flight space considered.

The estimation of $\hat{X}_{i}(\tau + \Delta\tau)$ naturally depends on the number of normal distributions involved within the superposition.
For the case of a single normal distribution, the search for the global optimum is trivial, as the mean coincides with the mode.
However, when multiple a mixture of normal distributions is taken into account, estimating $\hat{X}_{i}(\tau + \Delta\tau)$ analytically is generally infeasible, and computational methods are required to approximate the optimal value.

\subsection{Model's Architecture}
\begin{figure}
    \centering
    \includegraphics[width=\linewidth]
    {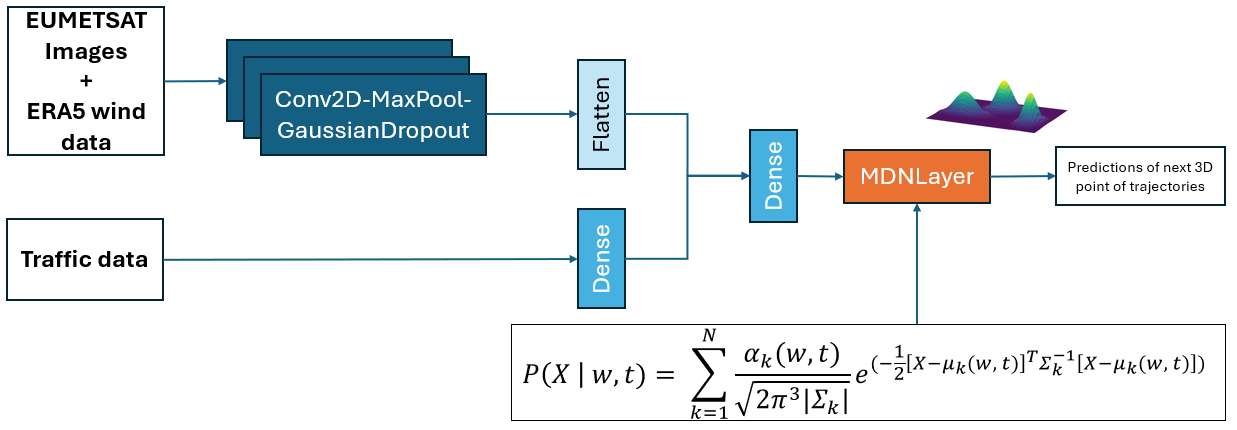}
    \caption{A schematic representation of the \gls{3dgmm}}
    \label{fig: scheme_model}
\end{figure}

The \gls{3dgmm} is devised to process both meteorological and traffic data simultaneously.
For this, we structured the model through a two-branch approach.
We can summarize the model's architecture through the following steps
\begin{enumerate}
    \item The meteorological data (convective weather and wind features) are processed through the first branch where the data are propagated through a combination of 2-D convolutional and max-pooling layers, powered by \emph{Rational Activation Functions} \cite[]{boulle2020rational}.
    This combination of convolutional and max-pooling layers was repeated three times.
    The scope of these operations is to capture salient patterns from the weather images while reducing their dimensionality.
    The latent description of data is then flattened through a \emph{Flatten Layer}.
    
    \item Air traffic data (traffic features) are processed through 3 Dense layers leading to a later description of the traffic data themselves

    \item Both the flattened latent descriptions coming from the two branches are then stacked together and then propagated through a further dense layer.
    
    \item At last, the \gls{mdn} Layer is applied.
    The \gls{mdn} layer estimates the Gaussian mixture parameters. 
    It consists of three-sub-branches dense layers, one estimating the parameters of the mixture (the condition \ref{eq: condition_prob_sum_1} is met by utilizing a softmax activation function), another one estimating the mean vector (no activation function, since the mean assume any real value), and the third one estimating the values of the covariance matrix.
\end{enumerate}
Note that the estimation of the covariance is achieved by constraining the matrix itself to be semi-definite positive and symmetric.
For in specific, we devised a custom layer with the scope of processing inputs through matrix operations to ensure that the covariance matrices derived in the model are Semi-Positive: this is achieved by computing a product of the input matrix with its transpose, thus yielding a symmetric result.
In addition, we ensured the computational stability of such a custom layer.
Similarly to the Tichonov regularization, we added a small constant to the diagonal entries of the covariance matrices.
Also, we opted for simplifying the construction of the covariance matrices by implementing the Cholesky decomposition.
This strategy allowed us to reconstruct the covariance matrix while maintaining its symmetric positive semi-definite property, critical for multivariate Gaussian distribution calculations in the MDN layer.
A summary scheme of the model is reported in figure \ref{fig: scheme_model}.
%%model image

\subsection{Model's Pre-processing}\label{sec: pre-processing}
Numerical-type traffic data and flight coordinates were pre-processed through a \emph{Power Transform}. Specifically, we utilized a Yeo–Johnson transform.
This transformation was employed to make the data distribution more Gaussian-like.
Additionally, it facilitated the rescaling of variables such as latitude, longitude, and altitude, which otherwise would have exhibited disproportionately large values for training the model effectively. 
Note that such a rescale preserves the consistency of \gls{3dgmm} accurate predictions, even when the data is reverted to its original scale prior to the power transformation.
This ensures that the transformations do not introduce distortions when returning the predictions to the original feature space; see appendix \ref{apx: power_transform}.

%Pre-Processing of Images. Walvelet Decoposition up to 95\% of energy
The \gls{3dgmm} is provided with 2-D grid-structured data (images) that includes weather-related information. 
These data were pre-processed using a 2-D wavelet transform, specifically the \emph{2-D Haar Wavelets}, for denoising and compression. 
To preserve 90\% of the energy in each image, we applied a criterion that involved removing the first wavelet components. 
Consequently, we used images reconstructed from the wavelet coefficients, excluding the first two levels of the transform.
More insights into this pre-processing step are discussed in the appendix \ref{apx: Wavelet Pre-processing}.

\subsection{Model's Selection}

This section outlines the methodology we employed to select the most performative architectures for the \gls{3dgmm}.
In particular, we shall focus our attention on large lead-time models.
An extensive hyperparameter optimization was conducted using a GridSearch approach. 
This method systematically evaluated combinations of hyperparameters, allowing fine-tuning of critical parameters to improve the model architecture and predictive capabilities. 
While searching for the best architecture, we accounted for several hyperparameters such as the \emph{number of filters in convolutional layers}, the \emph{kernel sizes}, the \emph{activation functions}, the \emph{depth of convolutional layers}, the \emph{depth of dense layers} within the traffic-related subnetwork, the \emph{Dropout rates}, the \emph{Mixture components} (i.e, the number of multi-variate normal distributions involved in the model), the \emph{learning rate}, and the \emph{batch size}.
We opted for a cross-validation strategy to select the optimal network configuration, identifying the model with the architecture that achieved the overall lowest \gls{mape} as the best choice.
Specifically, we utilized a 5-fold cross-validation scheme.
The best configurations for long lead times such as 30, 45, and 60minutes are shown in Table \ref{tab:optimal_configs_transposed}.

For all of these models, a single component of the mixture was sufficient to make accurate predictions. Several consistent patterns emerged across the optimal configurations. 
The Rational Activation Function (RAF) activation function \citep{boulle2020rational} proved particularly effective in capturing the complex patterns in the data and outperformed more traditional activation functions, such as \textit{tanh} and \textit{sigmoid}.
A depth of two convolutional layers for weather input and a single layer for traffic input provided sufficient representational capacity.
Smaller learning rates facilitated stable convergence during training, reducing the risk of oscillations in the loss function. Smaller batch sizes contributed to improved performance, likely due to more granular gradient updates in scenarios with higher variability.
%%%%
\begin{table}[h]
\centering
\begin{tabular}{|c|c|c|c|}
\hline
      & \textbf{30 min} & \textbf{45 min} & \textbf{60 min} \\ \hline
\textbf{Activation}     & RAF             & RAF             & RAF             \\ \hline
\textbf{Conv Layers}    & 2               & 2               & 2               \\ \hline
\textbf{Traffic Layer}  & 1               & 1               & 1               \\ \hline
\textbf{Dropout Rate}   & 0.25            & 0.25            & 0.5             \\ \hline
\textbf{Filter Size}    & 8               & 16              & 8               \\ \hline
\textbf{Kernel Size}    & 3               & 3               & 3               \\ \hline
\textbf{Batch Size}     & 64              & 32              & 32              \\ \hline
\end{tabular}
\caption{Optimal configurations for different long lead times.}
\label{tab:optimal_configs_transposed}
\end{table}

\section{Results}
\begin{figure}
    \centering
    \includegraphics[width=1\textwidth]{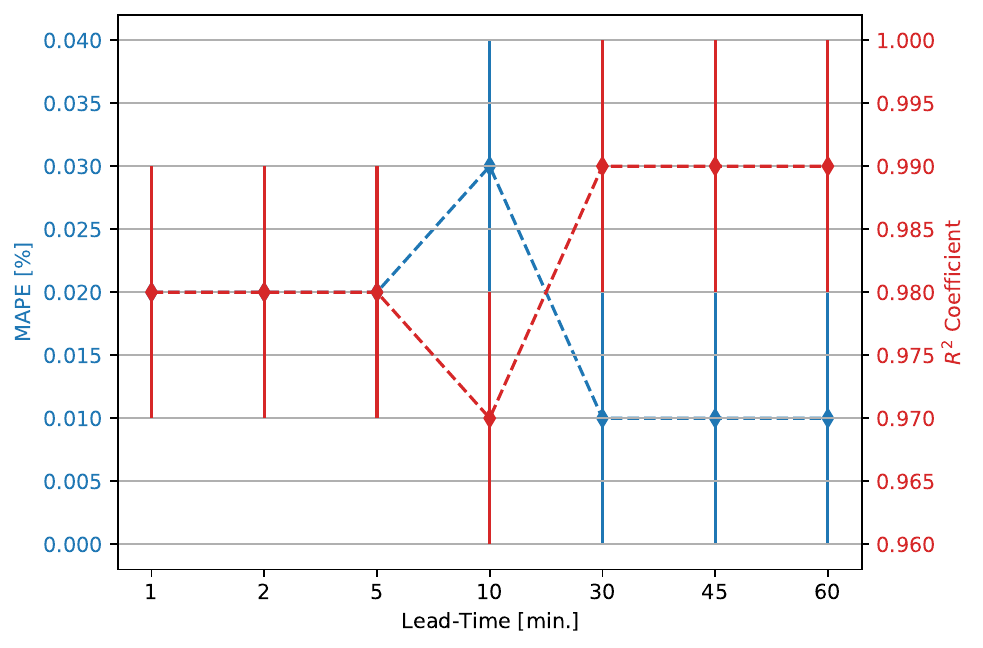}
    \caption{Double y-axis plot with the validation metrics for the \gls{3dgmm} at various lead-times.
    Red dots denote the average $R^{2}$ coefficient, while the blue ones denote the average \gls{mape}.
    Error bars were evaluated as the standard error mean.
    On the x-axis the values of the lead time, while along the vertical axis the values attained by the metrics.
    }
    \label{fig: Results_AI4ATM}
\end{figure}

To assess the goodness of the \gls{3dgmm}, we refer to two metrics; the \gls{mape} and the coefficient of determination (denoted $R^2$).
We evaluated these metrics on the predictions of the \emph{test set}.
We considered predictions at various lead times of forecasting; specifically, 1, 2, 5, 10, 30, 45, and 60 minutes.
Note that, we trained a specific \gls{3dgmm} model for each lead-time.
In figure \ref{fig: Results_AI4ATM}, we reported the metrics, for each lead time.
For each lead time, we observed very accurate predictions at both short and long lead times.
At a short lead time of 1-2 minutes, the \gls{mape} is $0.02 \pm 0.01$, while the $R^2$ is $0.98 \pm 0.01.$
At a large lead time of 60 minutes, we observed a \gls{mape} of $0.01 \pm 0.01$; the $R^2$ was equal to $0.99 \pm 0.01.$

When transporting the predictions at large lead times back to the original space, we obtained for the \gls{mape} the following coordinate-specific values: For the 30-minute lead times; altitude (above 1000 meters) $0.001, \pm 0.001$, longitude  $0.001, \pm 0.001$, and latitude $0.0001, \pm 0.0001.$
For a 45-minute lead time, altitude (above 1000 meters) $0.003, \pm 0.001$, longitude  $0.001, \pm 0.001$, and latitude $0.0003 \pm 0.0001.$
For a 60-minute lead time, altitude (above 1000 meters) $0.002, \pm 0.001$, longitude  $0.001, \pm 0.001$, and latitude $0.0001 \pm 0.0001.$

When evaluating a reference model, such as FlightBERT \citep{guo2022flightbert}, we observed high predictive accuracy across different lead times. 
For a 60-minute lead time, the model achieved a \gls{mape} of $0.021 \pm 0.001$ and an $R^2$ of $0.981 \pm 0.001$. 
At a 45-minute lead time, the \gls{mape} was $0.018 \pm 0.001$ with an $R^2$ of $0.982 \pm 0.001$, while for a 30-minute lead time, we recorded a \gls{mape} of $0.026 \pm 0.001$ and an $R^2$ of $0.973 \pm 0.001$. 

A comparison of the training times between the two models revealed that the \gls{3dgmm} trains significantly faster than FlightBERT. Specifically, using datasets of equal size, the \gls{3dgmm} completed training in approximately 4 hours, whereas FlightBERT required around 15 hours.

\section{Explaining the predictions}\label{sec: Explaining the predictions}
To explain the predictions of the \gls{3dgmm}, we employed the \gls{vg} algorithm \cite[]{simonyan2013deep}.
Our goal was to visually identify which input variables had the greatest impact on the \gls{3dgmm} predictions.
Among gradient-based \gls{xai} algorithms, the \gls{vg} method is one of the most simple and straightforward. 
Unlike other gradient-based methods (for instance, see \cite[]{sundararajan2017axiomatic, selvaraju2020grad}), \gls{vg} relies solely on this straightforward gradient evaluation, making it a simpler and more focused method.
The \gls{vg} determines feature importance exclusively by focusing on the change between the model's output and the individual input features, that is, through their gradients. 
According to this approach, the magnitude of a gradient serves as an indicator of the feature's importance; the larger the magnitude, the greater the feature's significance.
In the case of the \gls{3dgmm}, the \gls{vg} was utilized to compare gradients from both operational input branches that constitute the model. 

To make the explainability of predictions easier to understand, we constructed a \gls{vg}-based saliency map for each instance, following a structured methodology. 
For the traffic features, the magnitude of the gradients was utilized directly as an indicator of saliency, reflecting their individual contributions to the model's predictions. 
For weather data, we adopted a different approach: the saliency of each weather channel was determined by summing the magnitudes of the gradients across all pixels in the corresponding weather image, yielding a global saliency score. 
To further enhance interpretability, all gradient magnitudes were normalized so that their total equalled one. 
This normalization process allowed the saliency of each feature to be expressed as a rate between 0 and 1, providing a clear and intuitive representation of their relative importance in influencing the \gls{3dgmm} predictions.    
Thus, we evaluated the saliency map for each instance of the train test.
Then, we considered the \emph{average of the saliency maps} to obtain an overall level of features' importance;
error bars were estimated through the \gls{sem}.
Since the \gls{3dgmm} is based on the superposition of one only Multi-Variate Gaussian Distribution, the prediction mainly relies on the mean vectors.
For this reason, the \gls{vg} method was utilized to explain the outcomes of the nodes of the mean vector, i.e. the vector expressing the mean prediction for latitude, longitude, and altitude.
See appendix \ref{apx: Vanilla_Gradient} for a mathematical insight into this methodology.

\begin{figure}
    \centering
    \includegraphics[width= .8\textwidth]{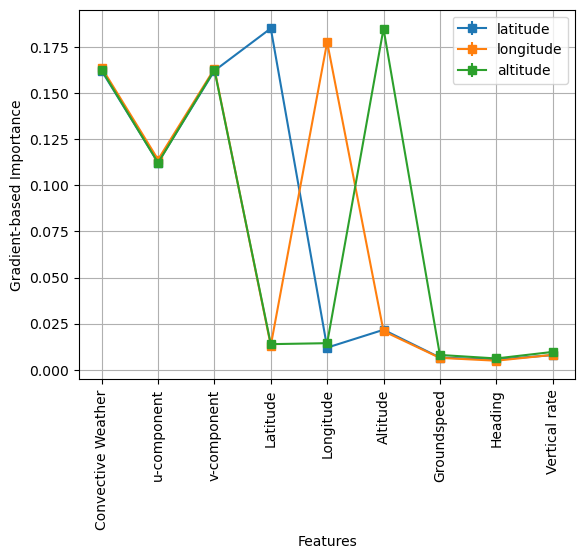}
    \caption{\gls{vg}-based saliency maps expressing the average percentage of importance explained by each feature.
    Each saliency map refers to one specific prediction attribute (i.e., latitude, longitude, and altitude).
    The importance is here expressed as the magnitude of the gradients.
    On the x-axis, the features are reported.
    On the y-axis, the estimated importance levels.
    Uncertainty, estimated using the \gls{sem}, is negligible and thus not visually discernible.
    }
    \label{fig: gradient_based_importance}
\end{figure}

In figure \ref{fig: gradient_based_importance}, the \emph{averaged saliency map} for the \gls{3dgmm} predictions is reported.
Note that a saliency map was constructed per each specific prediction attribute (i.e., latitude, longitude, and altitude). 
For convenience, we decided to show only the saliency maps for a 60-minute lead time model.
We recall that a total of 9 features were considered; see Table \ref{tab: complete_dataset}.% 
Notably, the saliency of these features is not uniform; their importance levels attain values above or below an idealized equal \emph{saliency benchmark} of about 0.11.
When comparing how features support the predictions, weather-related features generally exhibit higher saliency than traffic-related features. 
For instance, the features \emph{Convective Weather} and \emph{v-component} achieve importance levels as high as 0.16.
The feature \emph{u-component} shows a level of importance close to the saliency benchmark of 0.11. 
In contrast, features such as \emph{Ground speed}, \emph{Heading}, and \emph{Vertical rate}  demonstrate much lower importance, with saliency levels below 0.01.
Interestingly, positional features such as \emph{latitude}, \emph{longitude}, and \emph{altitude} stand out for their high saliency. 
Each of these features demonstrates significant individual contributions, with saliency values reaching up to 0.18, underscoring their critical roles in predicting their respective trajectories at future times.
Each of these features exhibits notable individual contributions, with saliency values peaking at 0.18, emphasizing their pivotal roles in forecasting their respective trajectories at future times. 
Specifically, this indicates that \emph{latitude at time} $t$ is a key factor in predicting \emph{latitude at a later time} $t+\tau$, \emph{longitude} plays an equally vital role in forecasting longitude at later times; likewise \emph{altitude} at a current time is indispensable for determining altitude in the future.
This suggests that \gls{3dgmm} predictions might keep track of the inherent dynamics of positional features.

\section{Conclusion}\label{sec: conclusion}
In this study, we proposed a \gls{3dgmm} for predicting long lead-time flight trajectory changes under adverse weather conditions. 
By leveraging high-resolution weather and traffic data, the model offers a comprehensive solution to forecast altitude, latitude, and longitude up to 60 minutes ahead. 
The results demonstrated exceptional predictive accuracy, with \gls{mape} values consistently low across both short and long lead times, confirming the effectiveness and robustness of the model. 

A key strength of this approach lies in its explainability, achieved through \gls{vg}-based saliency maps, which highlighted the critical role of positional and weather features. 
This not only enhances trust in the predictions but also provides valuable insights into the underlying dynamics of air traffic trajectories. 
Furthermore, the \gls{3dgmm} model exhibits superior computational efficiency compared to FlightBERT.
Specifically, training the \gls{3dgmm} requires approximately 4 hours, while FlightBERT demands around 15 hours on datasets of equivalent size.
This substantial reduction in training time enhances the model's scalability and adaptability for real-time operational settings. 

In addition to its efficiency, the \gls{3dgmm} consistently outperforms FlightBERT in terms of prediction accuracy across multiple lead times. For a 60-minute lead time, our model achieves a \gls{mape} of $0.01 \pm 0.01$ and an $R^2$ of $0.99 \pm 0.01$, surpassing FlightBERT’s corresponding values of $0.021 \pm 0.001$ and $0.981 \pm 0.001$. Similar improvements are observed for 45-minute and 30-minute lead times, reinforcing the reliability of the proposed method. 

The proposed \gls{3dgmm} model contributes to the evolving landscape of ATM by offering a scalable and interpretable framework capable of mitigating the impacts of weather-induced disruptions. Future work could explore integrating additional data sources, extending the model to broader airspaces, and refining the model’s capacity for real-time operational deployment.

\section*{Acknowledgment}
This research is supported by the Italian Ministry of Foreign Affairs and International Cooperation (MAECI) and the Agency for Science, Technology and Research (A*STAR), Singapore, under the First Executive Programme of Scientific and Technological Cooperation between Italy and Singapore for the years 2023–2025. Any opinions, findings, conclusions, or recommendations expressed in this material are those of the author(s) and do not necessarily reflect the views of the Italian Ministry of Foreign Affairs and International Cooperation or the Agency for Science, Technology and Research (A*STAR), Singapore.

\begin{appendices}
%\appendixpage
\noappendicestocpagenum
\addappheadtotoc

\section{Mathematical Details}

\subsection{Consistency of predictions}\label{apx: power_transform}

In this section, we aim to provide further insight into how the rescaling process impacts the consistency of \gls{3dgmm} predictions, particularly when the data is reverted to its original scale prior to the power transformation.
Let us suppose that $X_i$ represents the position of the $i$-th flight, i.e., altitude, longitude, and latitude.
Next, we denote by $\mathcal{T}(X_i)$ the 
Yeo-Johansen power transform.
We recall that the Yeo-Johansen transform is applied independently on altitude, longitude, and latitude.
For example, for the altitude $X^{(\text{altitude})}_{i}$ the power transform reads
\begin{equation}
    \mathcal{T}(X^{(\text{altitude})}_{i}|\lambda) = 
    \begin{cases}
        \frac{ \left(X^{(\text{altitude})}_{i}+1 \right)^{\lambda}-1}{\lambda} \quad \text{for }\lambda\neq 0;~X^{(\text{altitude})}_{i} \ge 0\\
        \log{\left(X^{(\text{altitude})}_{i}+1 \right)}\quad \text{for }\lambda = 0;~X^{(\text{altitude})}_{i} \ge 0\\
-\frac{ \left(-X^{(\text{altitude})}_{i}+1 \right)^{2-\lambda}-1}{2-\lambda} \quad \text{for }\lambda\neq 2;~X^{(\text{altitude})}_{i} < 0\\
        -\log{\left(-X^{(\text{altitude})}_{i}+1 \right)}\quad \text{for }\lambda = 2;~X^{(\text{altitude})}_{i} < 0;\\ 
    \end{cases}
\end{equation}
$\lambda$ is a parameter that is estimated
by optimizing the log-likelihood of the transformed data, with the goal of making the data as Gaussian-like as possible

As described, the predictions of the \gls{3dgmm} are based on the power-transformed values.
Let us suppose now that the \gls{3dgmm} prediction $\eta^{(\text{altitude})}_{i}$ (for the sake of illustration, we keep considering the altitude) is such that
\begin{equation}\notag
    \frac{|\eta^{(\text{altitude})}_{i}- \mathcal{T}(X^{(\text{altitude})}_{i}|\lambda)|}{\mathcal{T}(X^{(\text{altitude})}_{i})} < \epsilon_0;
\end{equation}
with $\epsilon_0$ any positive real-valued small quantity.
Note that, since the power transform is invertible, the prediction for the altitude can be written as
\begin{equation}
    \hat{X}^{(\text{altitude})}_{i} = \mathcal{T}^{-1}(\eta^{(\text{altitude})}_{i}|\lambda)
\end{equation}
Therefore, it is licit to write
\begin{equation}\label{eq: log_modulation_MAPE}
    \epsilon \left|X^{(\text{altitude})}_{i} 
    \frac{\partial \log{\mathcal{T}(Y^{(\text{altitude})}_{i}|\lambda)}}{\partial Y^{(\text{altitude})}_{i}} \ \bigg|_{Y = X^{(\text{altitude})}_{i}} \right|< \epsilon_0;
\end{equation}
where $\epsilon = \left| \frac{X^{(\text{altitude})}_{i} - \hat{X}^{(\text{altitude})}_{i}}{X^{(\text{altitude})}_{i}} \right|.$

As known, altitude, longitude and latitude take all positive values.
As a result, the evaluation of the modulation factor of \eqref{eq: log_modulation_MAPE}, can be restricted to the power-transforms utilizing positive-valued data.
That is,
\begin{equation}
    \begin{cases}
        \epsilon \left|X^{(\text{altitude}}_{i} \frac{\left(X^{(\text{altitude}}_{i} +1 \right)^{\lambda-1} }{ \left(X^{(\text{altitude}}_{i} +1 \right)^{\lambda} -1} \right| < \epsilon_0  \quad \text{for }\lambda\neq 0;~X^{(\text{altitude})}_{i} \ge 0 \\
\\
        \epsilon \left| \frac{X^{(\text{altitude}}_{i}}{ (\log{\left(X^{(\text{altitude}}_{i} +1\right)} \left(X^{(\text{altitude}}_{i} +1 \right)} \right| < \epsilon_0  \quad \text{for }\lambda =  0;~X^{(\text{altitude})}_{i} \ge 0 \\
        
    \end{cases}
\end{equation}

In the regime of large observations (i.e., $|X(^{\text{altitude})}| >> 1$), 
\begin{equation}\notag
    \begin{cases}
        \frac{\epsilon_0}{\epsilon} \gtrsim  1  \quad \text{for }\lambda\neq 0;~X^{(\text{altitude})}_{i} \ge 0 \\
\\
        \frac{\epsilon_0}{\epsilon} \gtrsim 0   \quad \text{for }\lambda =  0;~X^{(\text{altitude})}_{i} \ge 0 \\
        
    \end{cases}
\end{equation}
Thus, when the transform is not singular (i.e., $\lambda \neq 0$), the prediction of large observation is always more accurate in the original space (in the sense of \gls{mape}) rather than in the transformed space.

Note that the regime of large observations always applies to the observation under consideration.
Indeed, we selected altitudes at more than 3000 feet, while for latitude and longitude, values were in the range of $20^\circ$ to $90^{\circ}$. Additionally, we applied only non-singular power transformations. 
This shows the consistency of \gls{3dgmm} predictions, i.e., specifically, the \gls{mape}-estimated accuracy levels of \gls{3dgmm} can still be maintained when converting predictions back into the original space

\subsection{Wavelet Pre-processing}\label{apx: Wavelet Pre-processing}
Given a 2-D signal $X[i, j]$ (where the subscript $ij$ denotes the position $ij$) and an orthonormal set of 2-D wavelets, the corresponding 2-D wavelet decomposition of $X_{ij}$ consist of representing $X[i, j]$ as 
\begin{equation}
X[i, j] = \sum_{k, l} c_{LL}[k, l] \phi_{k, l}^{(Q)}(m, n) 
+ \sum_{k, l} c_{LH}[k, l] \psi_{k, l}^{\text{LH}}(m, n) 
+ \sum_{k, l} c_{HL}[k, l] \psi_{k, l}^{\text{HL}}(m, n) 
+ \sum_{k, l} c_{HH}[k, l] \psi_{k, l}^{\text{HH}}(m, n),
\end{equation}

where:

\begin{itemize}
    \item \( c_{LL}[k, l] \) denotes the \emph{approximation coefficients} at the lowest resolution (obtained using low-pass filters in both directions).
    \item \( c_{LH}[k, l] \) denotes the  \emph{detail coefficients} corresponding to horizontal details (low-pass in rows, high-pass in columns with).
    \item \( c_{HL}[k, l] \) denotes the \emph{detail coefficients} corresponding to vertical details (high-pass in rows, low-pass in columns).
    \item \( c_{HH}[k, l] \) denotes the \emph{detail coefficients} corresponding to diagonal details (high-pass in both directions).
\end{itemize}

The basis functions are defined as:
\begin{align*}
\phi_{k, l}(m, n) &= \phi(m - k) \phi(n - l), \\
\psi_{k, l}^{\text{LH}}(m, n) &= \phi(m - k) \psi(n - l), \\
\psi_{k, l}^{\text{HL}}(m, n) &= \psi(m - k) \phi(n - l), \\
\psi_{k, l}^{\text{HH}}(m, n) &= \psi(m - k) \psi(n - l);
\end{align*}
where \( \phi \) is the \emph{scaling function}, and \( \psi \) is the \emph{wavelet function}.
For example, in the \emph{db1 wavelet system} (i.e., Daubechies wavelet with one vanishing moment) the scaling function takes the form
\begin{equation}\notag
    \phi(x) =
\begin{cases} 
1, & 0 \leq x < 1, \\
0, & \text{otherwise}.
\end{cases}
\end{equation}
The wavelet function takes the form
\begin{equation}\notag
    \psi(x) =
\begin{cases} 
1, & 0 \leq x < \frac{1}{2}, \\
-1, & \frac{1}{2} \leq x < 1, \\
0, & \text{otherwise}.
\end{cases}
\end{equation}

If the wavelet transform is applied recursively, \( I[i, j] \) can be represented as:
\begin{equation}
\begin{split}
   I[i, j] =  \sum_{q=1}^Q & \left( 
\sum_{k, l} c_{LH}^{(j)}[k, l] \psi_{k, l}^{\text{LH},(j)}(m, n)  
+ \sum_{k, l} c_{HL}^{(j)}[k, l] \psi_{k, l}^{\text{HL},(j)}(m, n) 
+ \sum_{k, l} c_{HH}^{(j)}[k, l] \psi_{k, l}^{\text{HH},(j)}(m, n) 
\right)  + \\
& \hspace{8cm}+ \sum_{k, l} c_{LL}^{(Q)}[k, l] \phi_{k, l}^{(Q)}(m, n);
\end{split}
\end{equation}
where \( Q \) is the number of decomposition levels.

We recall, that the energy of the image $I[i, j]$ defined as
\begin{equation}
    \epsilon_{I} = \sum_{ij} I^{2}[i, j].
\end{equation}
In a Multi-level Wavelet Decomposition, the energy of a 2-D image is related to the wavelet coefficients by the formula
\begin{equation}
    \epsilon_{I} = \sum_{q=1}^Q \left( \sum_{m, n} |c_{LH}^{(q)}[m, n]|^2 + \sum_{m, n} |c_{HL}^{(q)}[m, n]|^2 + \sum_{m, n} |c_{HH}^{(q)}[m, n]|^2 \right) + \sum_{m, n} |c_{LL}^{(Q)}[m, n]|^2 .
\end{equation}
A well-designed wavelet transform concentrates most of the image's energy in the approximation coefficients $c_{LL}$, with smaller amounts in the detail coefficients $c_{LH}$,$c_{LH}$, $c_{HL}$, and $C_{HH}$. 
This property is useful in image compression, as the detail coefficients can be more aggressively quantized or thresholded without significantly impacting image quality, preserving the majority of the energy in the approximation coefficients.

\begin{figure}
    \centering
    \includegraphics[width=\linewidth]{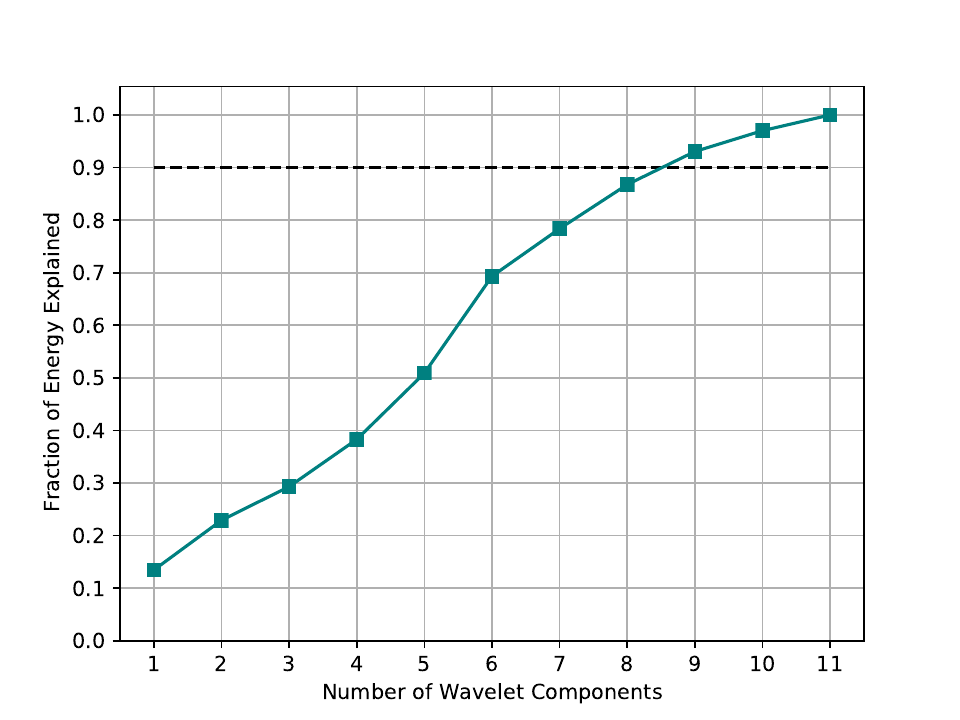}
    \caption{Average fraction of energy explained by reconstructing weather images from deepest wavelet level.
    95\% Confidence intervals were reported, but they are too small compared to the data points on the graph.
    The dotted line denotes the 90\% of energy explained.
    }
    \label{fig: wavelet_energy_explained}
\end{figure}

We utilized such a property to compress the weather data.
That is, starting from the most high-resolution coefficients, we dropped out the maximal number of wavelet levels, up to ensure a reconstruction, preserving at least 90\% of the energy.
Thus, we pre-processed the weather data by reconstructing imeges, through the db1 wavelet system, after discarding the first two wavelet levels; see \ref{fig: wavelet_energy_explained}.

\subsection{Vanilla Gradient-based Saliency Maps}\label{apx: Vanilla_Gradient}

%A sort of introduction...
Gradient-based algorithms assess how small changes in input features affect a neural network's output, providing insights into feature relevance via backpropagation.
These methods compute the gradient of the output with respect to the input, enabling the construction of saliency maps \cite[]{simonyan2013deep}, which outline the sensitivity of each input feature.

The \gls{vg} method, a simple and architecture-agnostic approach, uses Taylor expansion to approximate class score changes and identify features most influential to predictions. 
Saliency maps are derived by calculating gradients and normalizing their values for interpretability, allowing features' importance to be visualized on a standardized scale.

\subsubsection{Mathematical construction of saliency maps}

In this section, we provide a deeper mathematical insight into the construction of the \gls{vg}-based saliency maps, as previously introduced in Section \ref{sec: Explaining the predictions}.
We recall that the \gls{3dgmm} is based on the mixture of one Multi-variate Gaussian.
As a result, the \gls{3dgmm} predictions are essentially due to the mean vector.
Also, we recall that the Gaussian mean vector is estimated through a dense layer; the output nodes do not share any weight either on each other or with the others estimating other quantities such as the covariance matrix and the mixture weights.
This fact allows us to treat the \gls{3dgmm} prediction as an equivalent combination of three independent 1-D real outputs. 

%%%
Given these considerations, we can now pass to treat the mathematical construction of the saliency maps.
Let us consider the function $\Phi: \mathbf{R}^{N} \longrightarrow \mathbf{R}.$.
Given that \gls{3dgmm} predictions can be equivalently meant as a combination of three independent 1-D real outputs, we 
suppose for a moment that $\Phi$ is one equivalent function reproducing the prediction function learnt by the \gls{3dgmm} to predict one variable among altitude, longitude, and latitude.

The \gls{vg} determines the saliency of each input feature, based on the value of the gradient.
That is, the higher the absolute value of a variable's gradient, the more influence such a variable has on the model's predictions.
Given a small perturbation $\epsilon$, a variation of $\Phi$ can be written as
\begin{equation}\notag
    \Delta\Phi = \sum_{i = 1}^{N} \frac{\partial \Phi(\mathbf{x})}{\partial x_{i}} \epsilon.
\end{equation}

As introduced, the magnitude of the gradients determines how sensitive to a change in the inputs the function $\Phi$ is.
This is also reflected in the predictions.
Therefore, we can look at the quantity $ \left|\frac{\partial \Phi(\mathbf{x})}{\partial x_{i}}\right|$ as a measure of the importance of the $i$-th feature.

A single gradient explains one feature of an instance.
In addition, its absolute value does not indicate the feature's importance unless compared to gradients of other features.
To construct a saliency map for any 
specific instance of interest, we opted for normalizing the values of the gradients.
This choice can be motivated as an attribution of a  \emph{normalized portion of importance} to each feature relative to the others.
In formulae, the \emph{saliency map} for the generic $i$-th instance evaluated at the $j$-th feature an be expressed as 
\begin{equation}\notag
    \gamma_{ij} = \frac{| \frac{\partial \Phi(\mathbf{x}^{(i)})}{\partial x^{(i)}_{j}}|} {\sum_{k = 0}^{N} | \frac{\partial \Phi(\mathbf{x}^{(i)})}{\partial x^{(i)}_{k}}|} .  
\end{equation}
As a result, an overall attribution of the saliency of the generic $j$-th feature can be achieved by averaging all single-instance saliency maps, namely
\begin{equation}
    \Gamma_{j} = \frac{1}{M}\sum_{i = 0}^{M} \gamma_{i}.
\end{equation}
To estimate the error of such a measure one can use the \emph{Standard Error Mean}.

The strategy we have constructed remains valid as long as the input data are not grid-structured, e.g., they are arranged as a data matrix.
The \gls{3dgmm} have implemented, however, was conceived to get both batches of images and ordinary data matrices as inputs.
As a result, when considering a small variation $\Delta\Phi$, one should also take into account the gradient of each pixel $x_{ij}$ in the input images, i.e., $\frac{\Phi(\mathbf{x})}{x_{ij}}.$
The absolute value of each one of these gradients returns the particular piece of information about how important such a pixel is in supporting the final prediction.
In this context, normalizing all gradients to construct a saliency map would be redundant; a single channel of one image would give a sufficiently large number of attributions making the saliency impractical to use.
For this reason, we opted to construct the saliency maps by summing all the absolute gradients of an image channel.
The aggregation of the gradients through the sum can be justified as it reflects the propagation of errors across features. 
By summing the gradients, we account for how small changes in individual input features collectively influence the output, aligning with the principles of error propagation in systems with interdependent variables.

Hence, we constructed the saliency map for an instance propagated through the \gls{3dgmm} using absolute gradients. 
For input images, the absolute gradients of all pixels were aggregated to determine their overall importance. 
These saliency levels were then normalized to produce an instance-specific saliency map. 
Finally, an overall saliency map was obtained by averaging the normalized maps across instances.

\end{appendices}

\bibliography{references}
\bibliographystyle{apalike}

\end{document}